\documentclass[letterpaper, 10 pt, conference]{ieeeconf}  

\IEEEoverridecommandlockouts                              

\usepackage{amsmath} 
\usepackage{amssymb}  
\usepackage[hyphens]{url} 
\usepackage{comment} 
\usepackage{graphicx}
\usepackage{array}
\usepackage{pgfplots}
\usetikzlibrary{patterns}
\usepackage{subcaption}
\usepackage{forest} 
\usepackage{mathabx} 
\usepackage{multirow}

\definecolor{tumblue_dark}{HTML}{005293}
\definecolor{tumblue_medium}{HTML}{64A0C8}
\definecolor{tumblue_light}{HTML}{98C6EA}
\definecolor{tumgray}{HTML}{DAD7CB}
\definecolor{tumgreen}{HTML}{A2AD00}
\definecolor{tumorange}{HTML}{E37222}

\graphicspath{{figures/}}

\title{\LARGE \bf
A9-Dataset: Multi-Sensor Infrastructure-Based Dataset for Mobility Research
}

\author{Christian Cre{\ss}, Walter Zimmer, Leah Strand, Maximilian Fortkord,\\ Siyi Dai, Venkatnarayanan Lakshminarasimhan, Alois Knoll \\
\thanks{The authors are with the Chair of Robotics, Artificial Intelligence and Real-time Systems in the Department of Informatics, Technical University of Munich, Boltzmannstr. 3, 85748 Garching, Germany. Contact: {\tt\small christian.cress@tum.de, walter.zimmer@tum.de, leah.strand@tum.de, venkat.lakshmi@tum.de} }
\thanks{All authors contributed equally to this work.}
}

\begin{document}

\maketitle
\thispagestyle{empty}
\pagestyle{empty}


\begin{abstract}
	Data-intensive machine learning based techniques increasingly play a prominent role in the development of future mobility solutions - from driver assistance and automation functions in vehicles, to real-time traffic management systems realized through dedicated infrastructure. The availability of high quality real-world data is often an important prerequisite for the development and reliable deployment of such systems in large scale. Towards this endeavour, we present the A9-Dataset based on roadside sensor infrastructure from the 3 km long Providentia++ test field near Munich in Germany. The dataset includes anonymized and precision-timestamped multi-modal sensor and object data in high resolution, covering a variety of traffic situations. As part of the first set of data, which we describe in this paper, we provide camera and LiDAR frames from two overhead gantry bridges on the A9 autobahn with the corresponding objects labeled with 3D bounding boxes. The first set includes in total more than 1000 sensor frames and 14000 traffic objects. The dataset is available for download at \url{https://a9-dataset.com}.
\end{abstract}

\begin{keywords}
	Autonomous Driving, Mobility Research, Sensor Fusion, Artificial Intelligence, C-ITS

\end{keywords}


\section{Introduction}

Autonomous driving is becoming a key technology that already plays a pivotal role in the traffic sector. A major driver for this development is the winning streak of artificial intelligence (AI) in the 21st century. Current advances in the deep learning domain allow many perceptive tasks that were previously only solvable by humans to be assigned to machines. 
This is a prerequisite for self-driving cars that can autonomously perceive their environment and safely navigate through complex road systems. On top of that, it paves the way to future smart city applications and facilitates the transition to safe and secure transport. 

However, the development of these AI based mobility services is dependent on the availability of big data. The training of the sophisticated AI models requires an enormous quantity of labeled data. This is because the generalization ability of the models increases significantly with the variance in the training data. For instance, the appearance of the vehicles is substantially influenced by the viewing angle of the sensor. Therefore, customizing a perception system with training data from its particular perspective can give a huge performance boost.

\begin{figure}[htbp]
	\includegraphics[width=\linewidth]{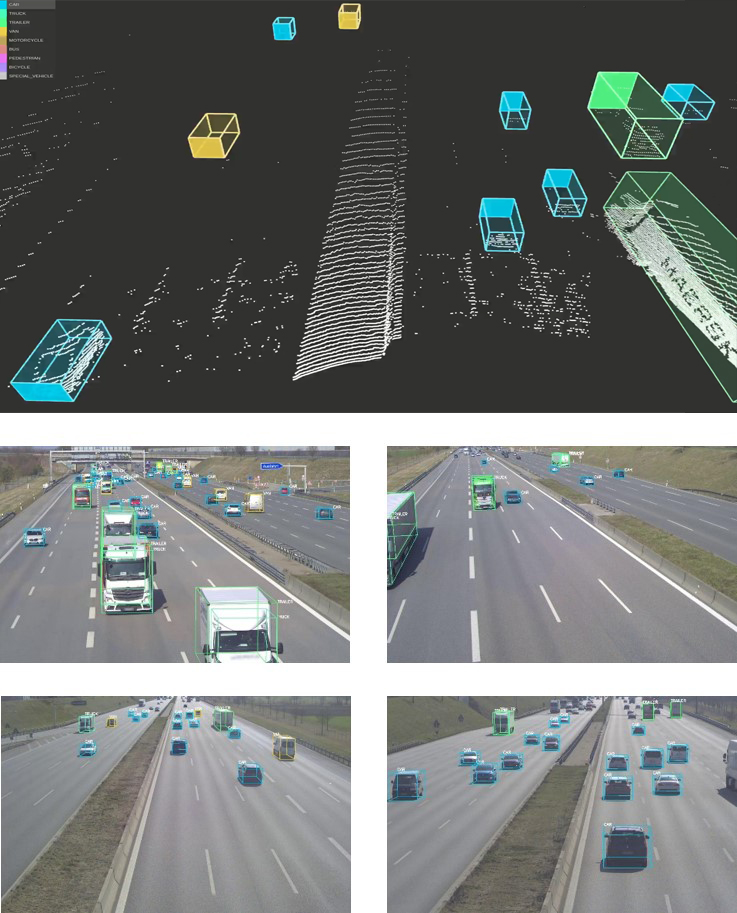}
	\caption{Labeled frames from LiDAR and cameras on two measurement stations on the freeway.}
	\label{fig:main_figure}
\end{figure}

To the best of our knowledge, there is no dataset publicly available that contains diverse road scenarios and was captured by stationary multi-modal sensors with a bird's eye view.
In particular, based on our experience, the supply of datasets aimed for the development of infrastructure-based perception systems is insufficient.

The A9-Dataset introduced in this paper aims to fill this gap. 
Here, we present the first release R0 which consists of 1098 labeled frames and 14,459 labeled 3D objects 
and was recorded from the sensors belonging to the hardware setup of the Providentia++ project (see Fig. \ref{fig:testbed_scheme}) close to Munich in Germany. 
The sensors are mounted to stationary infrastructure, namely gantry bridges and masts, and provide a far-reaching view of the road.
Due to the variety of sensors in our system, the dataset offers labeled images and LiDAR point clouds of multiple road segments and from different angles.
The test field covers diverse road types including the autobahn A9, rural roads and urban intersections. While the first release contains recordings of dense traffic on the autobahn, future releases will also contain the other sections. 
Since large-scale setups like the Providentia++ system are laborious and costly to construct, 
we want to share the data with the automotive perception research community.
It is intended for improving the robustness of vehicle detection models in general and for increasing the perception performance of intelligent infrastructure systems in particular.

\section{Related Work}
\label{sec:related_work}

There exist numerous autonomous driving datasets focusing on the on-board sensor data collection.
One of the earliest and most famous is the KITTI dataset \cite{KITTI} published almost a decade ago.
It has had a great impact on autonomous driving research with more than 13,000 academic citations since its release. 
Apart from that, Lyft Level 5 \cite{houston2020one} is one of the largest autonomous driving datasets in the industry containing over 1,000 hours of data and 170,000 scenes. 
Meanwhile, being considered as the first dataset to carry the full autonomous vehicle sensor suite, nuScenes \cite{nuscenes} comprises 1,000 scenes annotated with 3D bounding boxes. It has 7 times the number of annotations and 100 times the number of images as the pioneering KITTI dataset.
The Argoverse open-source dataset \cite{argoverse} is dedicated to 3D tracking and motion forecasting. It comprises 3D tracking annotations for 113 scenes and additionally 324,557 interesting vehicle trajectories extracted from over 1,000 driving hours. 
In addition, the Waymo open dataset was recently released. It is organized into a perception dataset \cite{waymo} with over 100,000 scenes and a motion dataset \cite{ettinger2021large} with 1,150 scenes containing 2D and 3D annotations captured across a range of urban and suburban roads.

Data recorded from an aerial perspective is considered to be valuable for fostering autonomous driving and traffic simulation research.
With this intention, the Next Generation SIMulation (NGSIM) dataset \cite{US101,I80} was released in 2006.
Here, the traffic was recorded by cameras mounted to surrounding buildings and processed to naturalistic vehicle trajectories.
However, many errors concerning vehicle dynamics have been found in the data \cite{coifman2017critical}, all of which harm the trust in analyses based on it. 
Furthermore, since 2005, camera-equipped drones have been used for aerial traffic monitoring \cite{puri2005survey}. 
The highD dataset \cite{krajewski2018highd} is considered to be the first public traffic dataset recorded by camera drones.
The large-scale dataset was recorded at six different locations of the German autobahn and contains trajectories from 110,500 vehicles.
It was recently supplemented by the datasets inD \cite{inD} and rounD \cite{rounD} that offer data from intersections and roundabouts, respectively.

While especially the highD dataset lays a sound groundwork for developments concerning aerial traffic monitoring, we want to complement the available data.
The A9-Dataset not only extends the data with the unique perspective of stationary infrastructure-mounted sensors, but moreover, our multi-modal sensor setup allows us to provide diverse and rich data of the traffic.

\section{Sensor Setup}
We recorded the A9-Dataset on the test bed Providentia++. This section gives an insight into the test bed as well as the sensors that we used.

\begin{figure}[hbtp]
	\includegraphics[width=\linewidth]{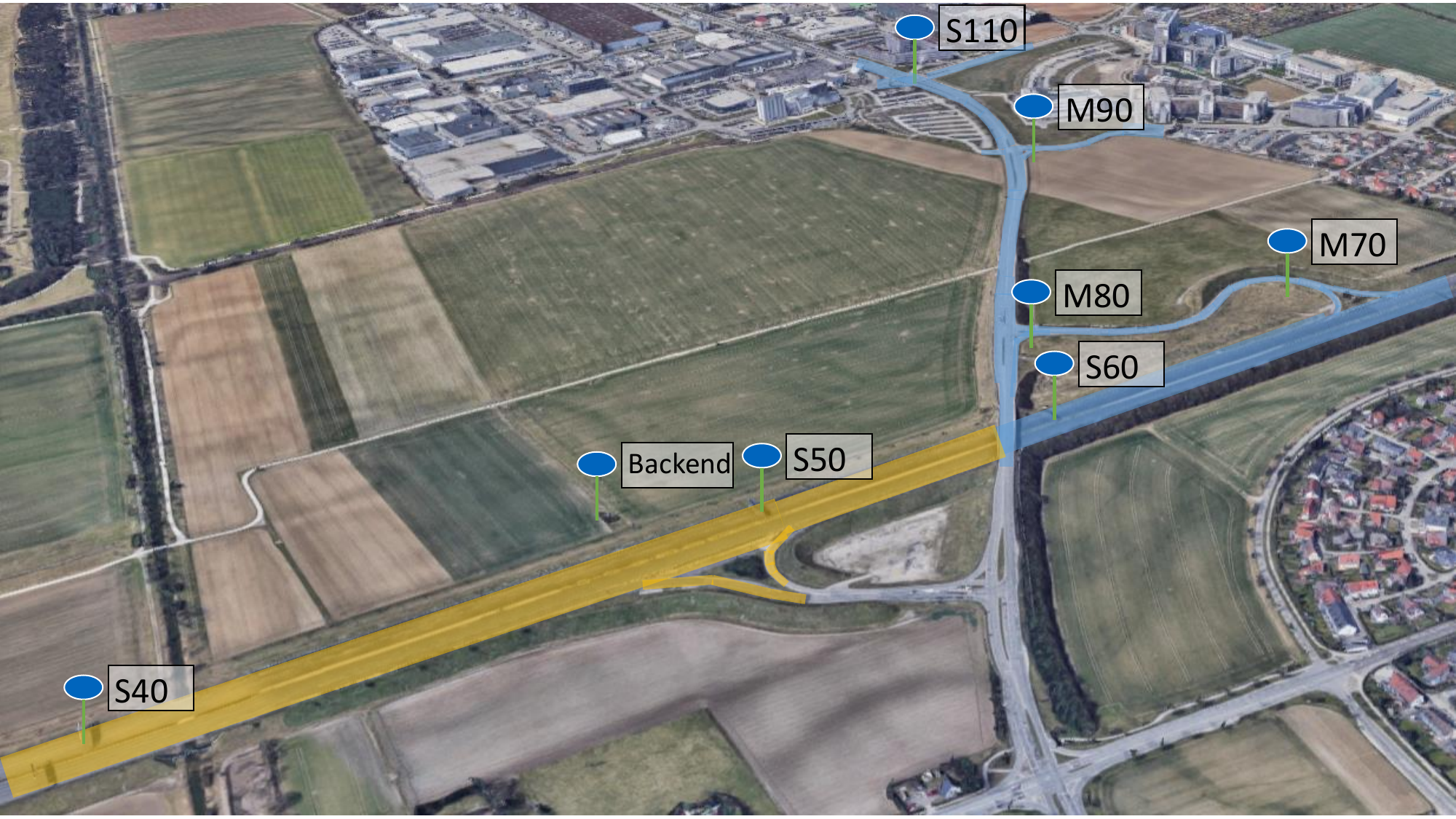}
	\caption{Overview of the test bed Providentia++ (Graphics made with Google).}
	\label{fig:testbed_scheme}
\end{figure}

\subsection{Test bed}
\label{sec:test_bed}
Based on the real world Intelligent Transportation System (ITS) Providentia \cite{krammer.2019.providentia,hinz.2017.providentia,lakshminarasimhan.2018.providentia}, we have extended the test bed into urban areas in the follow-up project Providentia++. On a total length of 3.5 km, the ITS is located at the autobahn A9 as well as the highway B471 near Munich. Therefore, the ITS covers with 7 measurement points the complete range of possible traffic scenarios: freeway, highway, roundabout and intersection in urban areas with pedestrians and bicycles. The main purpose of the ITS is the creation of digital twins of all traffic participants in real-time. Figure \ref{fig:testbed_scheme} shows a complete overview of the Providentia++ test bed. 

\begin{figure}[htbp]
	\includegraphics[width=\linewidth]{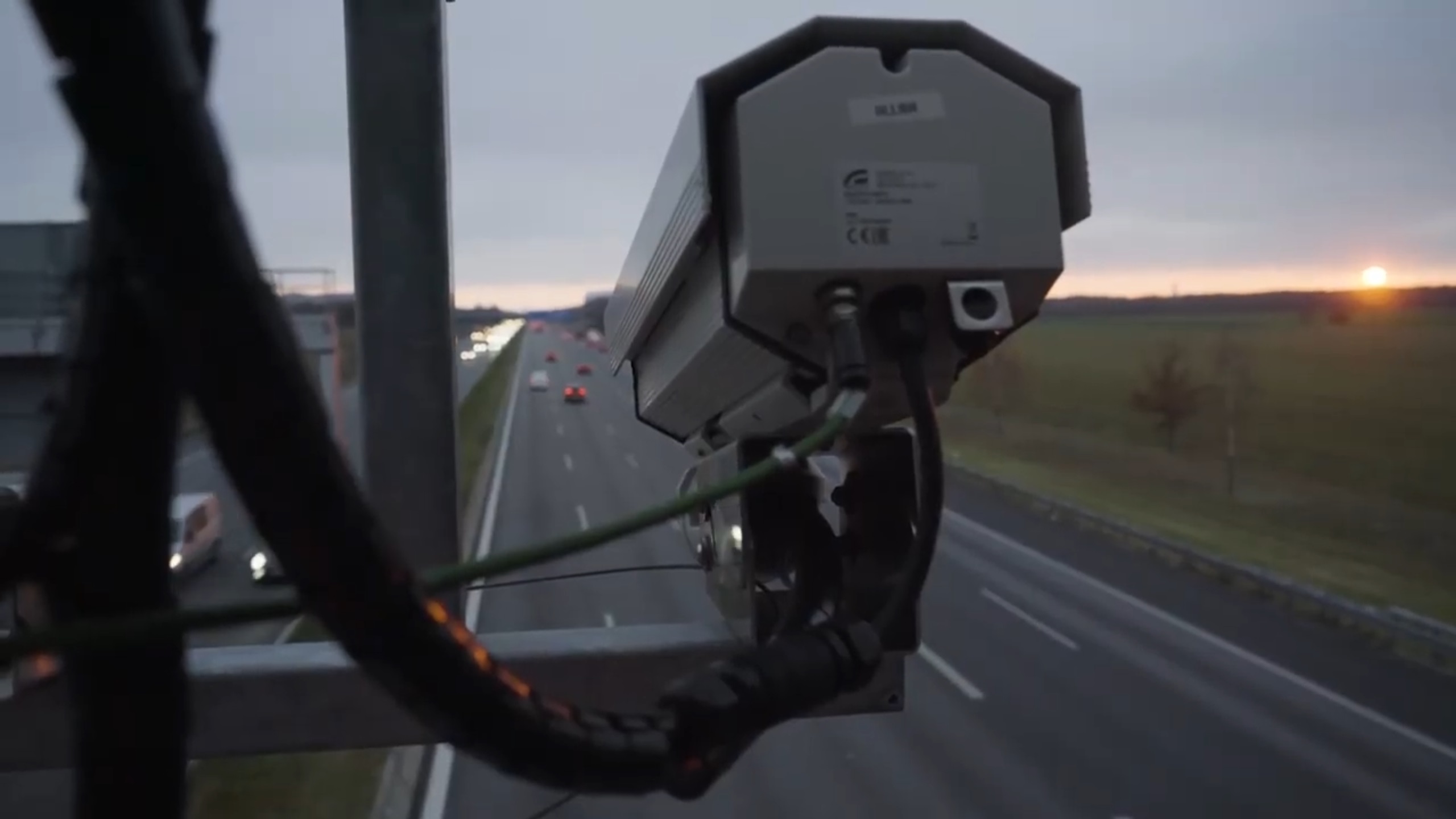}
	\caption{Camera perspective of the S50 camera in southern direction.}
	\label{fig:s50_camera}
\end{figure}

The presented dataset is based on the sensors of the measurement points S40 and S50 mounted on overhead gantry bridges along the A9 autobahn. The S40 sensor station contains one near field camera with 16 mm focal length, one far field camera with 50 mm focal length, and two Radars in north and south direction. Additionally, the S40 has two event-based cameras in north direction. On sensor station S50, the same camera and radar setup as on S40 is installed. Furthermore, S50 has two event-based cameras and two Valeo LiDARs in south direction. In the A9-Dataset, we have used the cameras of S40 north and S50 south as well as a temporary installed Ouster LiDAR on S50. The sensors capture the traffic from a bird's eyes view from a height of approximately 7 m. Figure \ref{fig:s50_camera} shows exemplary the perspective of the camera.    

\subsection{Sensors}
 
The A9-Dataset was recorded with the following sensors:\\
\textbf{Cameras:} 4x Basler ace acA1920-50gc, 1920x1200, Sony IMX174, glo. shutter, 1/1.2'', CMOS, color, GigE with 16/50 mm lenses.\\ 
\textbf{LiDAR:} 1x Ouster OS1-64 (gen 2), 64 vert. layers, 0.26-0.52\textdegree~vert. ang. res., 0.18-0.7\textdegree~horiz. ang. res., 33.1\textdegree~vert. FOV, 360\textdegree~horiz. FOV, 120 m range, 1.5-10 cm accuracy, 131,072 points per scan, 1,310,720 points per sec.
\section{Dataset Description}
In this section, we present the content and the format of the dataset in more detail.
Particularly, we focus on the first release R0 that comprises camera and LiDAR recordings of dense traffic on the A9 autobahn during daylight.
The data has a total size of 3.85 GB.

\subsection{Subsets}
Our dataset is organized into four subsets S1 to S4, whereby the first two subsets contain the camera data and the latter two the LiDAR scans.
More specifically, the first subset S1 was created by randomly sampling the image streams of 4 different cameras equipped with two types of lenses.
They are mounted on the two overhead gantry bridges denoted as S40 and S50 (see Fig. \ref{fig:testbed_scheme}). 
By contrast, the second subset S2 shows a traffic sequence observed by one camera that is located at the sensor station S40. 
The two LiDAR subsets S3 and S4 contain two different traffic sequences recorded by one LiDAR sensor on the gantry bridge S50.
The contents and specifics of the four subsets are listed in Table \ref{tbl:subsets}.

\begin{table*}[h]
	\caption{Details on the four subsets belonging to the dataset release R0.}
	\label{tbl:subsets}
	\centering
	\begin{tabular}{|p{3.2cm}||p{2.3cm}||p{2.3cm}||p{2.3cm}||p{2.3cm}||p{2.3cm}|}
		\hline
		& R0\_S1   & R0\_S2   & R0\_S3   & R0\_S4   & R0\_total  \\
		\hline
		time of recording & winter + summer& summer & winter & summer & \\
		\hline
		sensors & 4 cameras & 1 camera & 1 LiDAR & 1 LiDAR & \\
		\hline
		data type &	images & images & point clouds & point clouds & \\
		\hline
		temporal properties & random frames & 24s sequence & 60s sequence  & 122s sequence & \\
		& & at 2.5Hz & at 2.5Hz & at 2.5Hz & \\
		\hline
		number of frames             & 582  & 60   & 150  & 306  & 1,098 \\
		\hline
		number of labels             & 9,331 & 2,024 & 1,585 & 1,519 & 14,459 \\ 
		\hline
		average labels per frame  & 16   & 33   & 11   & 5    & 13.17 \\
		\hline
	\end{tabular}
\end{table*}

\subsection{Data Format}
\label{sec:data_format}

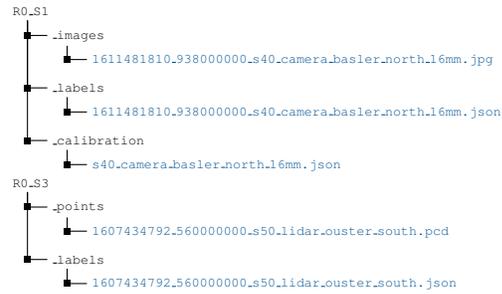
\begin{figure}[h]
	\begin{subfigure}[t]{\linewidth}
		{\tiny
			\begin{forest}
				for tree={
					font=\ttfamily,
					grow'=0,
					child anchor=west,
					parent anchor=south,
					anchor=west,
					calign=first,
					edge path={
						\noexpand\path [draw, \forestoption{edge}]
						(!u.south west) +(7.5pt,0) |- node[fill,inner sep=1.25pt] {} (.child anchor)\forestoption{edge label};
					},
					before typesetting nodes={
						if n=1
						{insert before={[,phantom]}}
						{}
					},
					fit=band,
					before computing xy={l=15pt},
				}
				[R0\_S1
				[\_images
				[{\color{tumblue_dark}\texttt{1611481810\_938000000\_s40\_camera\_basler\_north\_16mm.jpg}}]
				]
				[\_labels
				[{\color{tumblue_dark}\texttt{1611481810\_938000000\_s40\_camera\_basler\_north\_16mm.json}}]
				]
				[\_calibration
				[{\color{tumblue_dark}\texttt{s40\_camera\_basler\_north\_16mm.json}}]
				]
				]
			\end{forest}
		}
	\end{subfigure}
	\begin{subfigure}[t]{\linewidth}
		{\tiny
			\begin{forest}
				for tree={
					font=\ttfamily,
					grow'=0,
					child anchor=west,
					parent anchor=south,
					anchor=west,
					calign=first,
					edge path={
						\noexpand\path [draw, \forestoption{edge}]
						(!u.south west) +(7.5pt,0) |- node[fill,inner sep=1.25pt] {} (.child anchor)\forestoption{edge label};
					},
					before typesetting nodes={
						if n=1
						{insert before={[,phantom]}}
						{}
					},
					fit=band,
					before computing xy={l=15pt},
				}
				[R0\_S3
				[\_points
				[{\color{tumblue_dark}\texttt{1607434792\_560000000\_s50\_lidar\_ouster\_south.pcd}}]
				]
				[\_labels
				[{\color{tumblue_dark}\texttt{1607434792\_560000000\_s50\_lidar\_ouster\_south.json}}]
				]
				]
			\end{forest}
		}
	\end{subfigure}
	\caption{Structure of camera and LiDAR sets (S1/S3).}
	\label{fig:folder_structure_r0_s3}
\end{figure}

The structure of both sets is displayed in Fig. \ref{fig:folder_structure_r0_s3}. Camera images and their corresponding label files are located in the first two sub sets (S1 and S2). They have the same file name that includes the timestamp in seconds and nano seconds and the sensor ID which is made up of the sensor station, the sensor type, sensor manufacturer, sensor direction and focal length of the lens. Furthermore, we provide intrinsic and extrinsic calibration matrices for the cameras. Labels and point clouds are part of the LiDAR set S3 and S4. Corresponding files in the LiDAR dataset have the same file name, which include the timestamp and the sensor ID. 

The content of the dataset is formatted as follows: \\
\textbf{Image data:} RGB image data is stored in JPG format (with a file size of ~200 KB) to optimize storage resources. The data was recorded with Basler cameras. The images have a resolution of 1920x1200 pixels.\\
\textbf{Point cloud data:} Point cloud scans were extracted into PCD (point cloud data) ASCII files where one file represents a single point cloud scan. An Ouster LiDAR file represents a full-surround scan. The PCD file contains the following attributes for each LiDAR point: x, y, z, intensity, timestamp, reflectivity, ring (layer), ambient and range value.\\
\textbf{Calibration data:} Sensor calibration data is provided in json format for each camera sensor. It contains the image width and height, the camera intrinsic information, the extrinsic calibration data and the projection matrix. \\
\textbf{Labels:} All sensor data was labeled in the json format using an extended version of the 3D Bounding Box Annotation Tool (3D BAT) \cite{zimmer20193d}. One json label file was created for each frame. It contains the corresponding image file name, a timestamp, the weather type, and the object labels. Each object contains a unique ID within the frame, a classification category, cuboid information and attributes. The cuboid information is given as location (in meters), dimension (in meters) and orientation (in degree). The location of the objects in the sets S1 and S2 is given in a locally defined coordinate frame on road-level. Its origin is set to the GPS position 48.241537 11.639538 and it is oriented following the freeway in the direction of south. Projected 3D cuboid coordinates are also given within the attributes of each object within the S1 and S2 set. They describe the 8 corner positions of the cuboid and are given in pixel coordinates. S3 and S4 sets contain the location in the sensor coordinate frame of the LiDAR. Attributes are different for each class. Cars, trucks, vans and busses were labeled with the number of trailers. Trailers have a type (container, tanker,box, dump, vehicle transporter, mixer, flatbed or other). Trucks have a truck type (pickup truck or delivery truck). Busses can be of type rigid or bendy. Motorcycles and bicycles can be tagged with the attribute "has rider". Motorcycles also have a type attribute (motorbike, scooter). Other vehicles have a flashing lights attribute. 

\subsection{A9 Development Kit}
The A9 Development Kit (a9-dev-kit)\footnote{\url{https://github.com/providentia-project/a9-dev-kit}} provides a dataset loader for images, point clouds, labels and calibration files. It contains a parser to read json label files and a preprocessing module to reduce e.g. noise in point cloud scans or to undistort camera images. Furthermore, some data filtering scripts allow you to filter for specific classes or for labels that contain at least 20 points inside a 3D box. A loader for calibration files is also provided to load the intrinsic and extrinsic calibration information. The projection matrix is used to visualize the 2D and 3D labels on cameras images and LiDAR point clouds. In addition, a data converter/exporter enables you to convert the labels from OpenLABEL format into other formats like KITTI, COCO or YOLO. Finally, a model evaluation script is provided to benchmark your models on the A9-Dataset. The README file in the linked repository provides more information about the usage of the dataset with the development kit.

\subsection{Statistics}
\label{sec:statistics}
We provide several figures and tables showcasing the wide range of data that has been collected on the test stretch. Table \ref{tbl:object_classes} shows the labeled object classes. Here, we would like to note that some classes are underrepresented, e.g. motorcycles or bicycles. Furthermore, our dataset in this version has limited variance in terms of lighting conditions. It mainly contains images in good visibility conditions during the day. In future versions of our data set, we will take additional visibility conditions into account. 

\begin{table}[htbp]
	\caption{Labeled object classes, the total amount of labels, average dimensions and average number of 3D LiDAR points per object among all 4 sets.}
	\label{tbl:object_classes}
	\centering
		\begin{tabular}{|p{1.3cm}||l||l||l||l||p{1.3cm}|}
		\hline
		Class & Labels & Length & Width & Height & 3D Points\\
		\hline
		Car        & 9,310 & 4.58  & 1.94 & 1.48 & 90\\
		\hline
		Trailer    & 1,828 & 12.49 & 2.82 & 3.79 & 1097\\
		\hline
		Truck      & 1,633 & 5.80  & 2.51 & 3.55 & 143\\
		\hline
		Van        & 1,366 & 5.68  & 2.09 & 2.22 & 195\\
		\hline
		Pedestrian & 220   & 0.55  & 0.49 & 1.68 & 42\\
		\hline
		Bus        & 59    & 13.58 & 2.61 & 3.35 & -\\
		\hline
		Motorcycle & 33    & 2.32  & 0.75 & 1.64 & -\\
		\hline
		Other      & 9     & 7.28  & 2.36 & 2.87 & -\\
		\hline
		Bicycle    & 1     & 1.42  & 0.48 & 1.24 & - \\
		\hline
	\end{tabular}
\end{table}

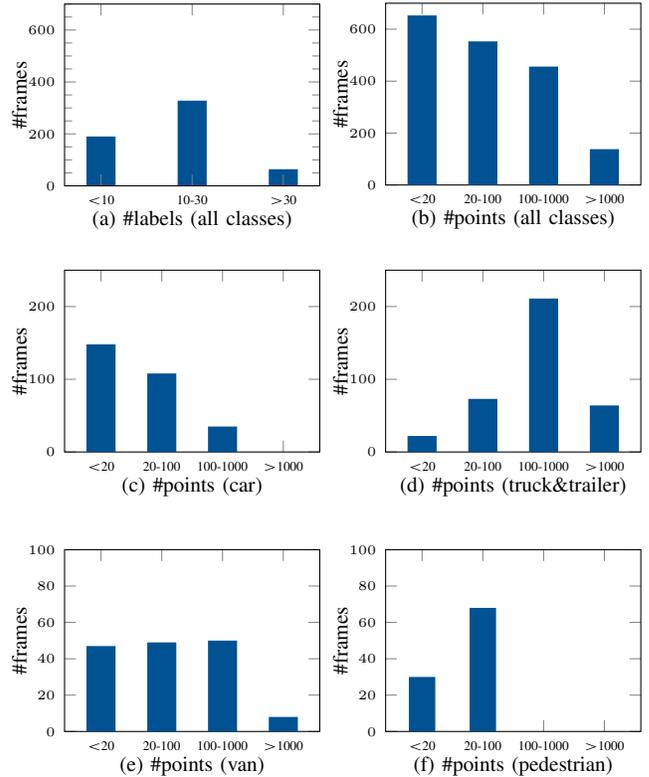
\begin{figure}
	\centering
	\begin{subfigure}[b]{0.48\linewidth}
		\centering
		\begin{tikzpicture}[font=\footnotesize]
			\begin{axis}[
				bar width=11pt,
				ymin=0, ymax=55,
				minor y tick num = 3,
				area style,
				height=40mm,
				width=1.2\linewidth,
				enlarge y limits = 0.0,
				enlarge x limits = 0.2,
				ymax=700,
				xlabel={(a) \#labels (all classes)},
				ylabel={\#frames},
				ylabel style={yshift=-4ex},
				xlabel style={yshift=2ex},
				x tick label style={font=\tiny},
				y tick label style={font=\tiny},
				symbolic x coords={$<$10,10-30,$>$30}
				]
				\addplot+[ybar,fill=tumblue_dark,draw=none] plot coordinates
				{($<$10, 190)
					(10-30, 328)
					($>$30, 64)};
			\end{axis}
		\end{tikzpicture}
		\label{fig:num_labels}
	\end{subfigure}
	\begin{subfigure}[b]{0.48\linewidth}   
		\centering 
		\begin{tikzpicture}[font=\footnotesize]
			\begin{axis}[
				ybar stacked,
				bar width=11pt,
				height=40mm,
				width=1.2\linewidth,
				enlarge y limits = 0.0,
				enlarge x limits = 0.2,
				ymin=0,
				ymax=700, 
				xlabel={(b) \#points (all classes)},
				ylabel={\#frames},
				ylabel style={yshift=-4ex},
				xlabel style={yshift=2ex},
				x tick label style={font=\tiny},
				y tick label style={font=\tiny},
				symbolic x coords={$<$20,20-100,100-1000,$>$1000},
				]
				\addplot+[ybar,fill=tumblue_dark,draw=none] plot coordinates {
					($<$20, 653) 
					(20-100, 553) 
					(100-1000, 456) 
					($>$1000, 138) 
				};
			\end{axis}
		\end{tikzpicture}
		\label{fig:num_points_objects}
	\end{subfigure}
	\begin{subfigure}[b]{0.48\linewidth}  
		\centering
		\begin{tikzpicture}[font=\footnotesize]
			\begin{axis}[
				ybar stacked,
				bar width=11pt,
				height=40mm,
				width=1.2\linewidth,
				enlarge y limits = 0.0,
				enlarge x limits = 0.2,
				ymin=0,
				ymax=250, 
				xlabel={(c) \#points (car)},
				ylabel={\#frames},
				ylabel style={yshift=-4ex},
				xlabel style={yshift=2ex},
				x tick label style={font=\tiny},
				y tick label style={font=\tiny},
				symbolic x coords={$<$20,20-100,100-1000,$>$1000},
				]
				\addplot+[ybar,fill=tumblue_dark,draw=none] plot coordinates {
					($<$20, 148)
					(20-100, 108)
					(100-1000, 35)
					($>$1000, 0)
				};
			\end{axis}
		\end{tikzpicture}
		\label{fig:num_points_car}
	\end{subfigure}
	\begin{subfigure}[b]{0.48\linewidth}   
		\centering
		\begin{tikzpicture}[font=\footnotesize]
			\begin{axis}[
				ybar stacked,
				bar width=11pt,
				height=40mm,
				width=1.2\linewidth,
				enlarge y limits = 0.0,
				enlarge x limits = 0.2,
				ymin=0,
				ymax=250, 
				xlabel={(d) \#points (truck\&trailer)},
				ylabel={\#frames},
				ylabel style={yshift=-4ex},
				xlabel style={yshift=2ex},
				x tick label style={font=\tiny},
				y tick label style={font=\tiny},
				symbolic x coords={$<$20,20-100,100-1000,$>$1000},
				]
				\addplot+[ybar,fill=tumblue_dark,draw=none] plot coordinates {
					($<$20, 22)
					(20-100, 73)
					(100-1000, 211)
					($>$1000, 64)};
			\end{axis}
		\end{tikzpicture}
		\label{fig:num_points_trailer}
	\end{subfigure}
	\begin{subfigure}[b]{0.48\linewidth}   
		\centering 
		\begin{tikzpicture}[font=\footnotesize]
			\begin{axis}[
				ybar stacked,
				bar width=11pt,
				height=40mm,
				width=1.2\linewidth,
				enlarge y limits = 0.0,
				enlarge x limits = 0.2,
				ymin=0,
				ymax=100, 
				xlabel={(e) \#points (van)},
				ylabel={\#frames},
				ylabel style={yshift=-4ex},
				xlabel style={yshift=2ex},
				x tick label style={font=\tiny},
				y tick label style={font=\tiny},
				symbolic x coords={$<$20,20-100,100-1000,$>$1000},
				]
				\addplot+[ybar,fill=tumblue_dark,draw=none] plot coordinates {
					($<$20, 47)
					(20-100, 49)
					(100-1000, 50)
					($>$1000, 8)};
			\end{axis}
		\end{tikzpicture}
		\label{fig:num_points_van}
	\end{subfigure}
	\label{fig:camera_images}
	\begin{subfigure}[b]{0.48\linewidth}   
		\centering 
		\begin{tikzpicture}[font=\footnotesize]
			\begin{axis}[
				ybar stacked,
				height=40mm,
				width=1.2\linewidth,
				enlarge y limits = 0.0,
				enlarge x limits = 0.2,
				ymin=0,
				ymax=100, 
				xlabel={(f) \#points (pedestrian)},
				ylabel={\#frames},
				ylabel style={yshift=-4ex},
				xlabel style={yshift=2ex},
				x tick label style={font=\tiny},
				y tick label style={font=\tiny},
				symbolic x coords={$<$20,20-100,100-1000,$>$1000},
				]
				\addplot+[ybar,fill=tumblue_dark,draw=none] plot coordinates {
					($<$20, 30)
					(20-100, 68)
					(100-1000, 0)
					($>$1000, 0)
				};
			\end{axis}
		\end{tikzpicture}
		\label{fig:num_points_pedestrian}
	\end{subfigure}
	
	\caption{Statistics for labeled objects visualized as histograms. (a) Distribution of the number of labels within all frames of the dataset. (b) Distribution of the number of LiDAR points per object for (b) all classes, (c) cars, (d) trucks and trailers, (e) vans and (f) pedestrians.} 
	\label{fig:statistics_labeled_objects}
\end{figure}

Figure \ref{fig:statistics_labeled_objects} shows the distribution of number of labels per frame. Most of the frames contain between 10 and 30 labels, thereby indicating significant traffic density.
 
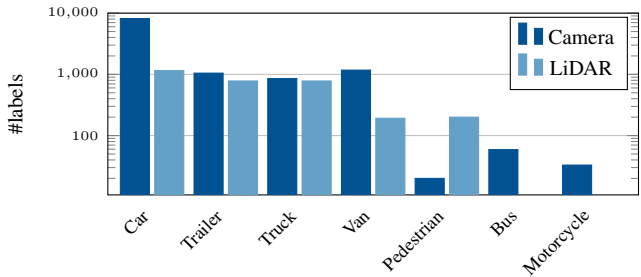
\begin{figure}
	\centering
	\begin{tikzpicture}[font=\footnotesize]
		\begin{axis}[					
			width  = 1.0\linewidth,
			height = 4cm,
			major x tick style = transparent,
			ybar,
			bar width=11pt,
			ymajorgrids = true,
			ylabel = {Run time speed},
			symbolic x coords={Car,Trailer, Truck, Van, Pedestrian, Bus, Motorcycle},
			xtick = data,
			scaled y ticks = false,
			ymax = 10000,
			ymode=log,
			log ticks with fixed point,
			xtick=data,
			x tick label style={rotate=45,anchor=east,font=\scriptsize},
			y tick label style={font=\tiny},
			ylabel={\#labels},
			]
			
			\addplot[style={tumblue_dark,fill=tumblue_dark,mark=none}]
			coordinates {(Car, 8158.0) (Trailer,1048.0) (Truck,853.0) (Van,1174.0) (Pedestrian,20) (Bus,59) (Motorcycle,33)};
			
			\addplot[style={tumblue_medium,fill=tumblue_medium,mark=none}]
			coordinates {(Car,1152.0) (Trailer,780.0) (Truck,780.0) (Van,192.0) (Pedestrian,200.0) (Bus,0) (Motorcycle,0) };
			
			\legend{Camera,LiDAR}
		\end{axis}
	\end{tikzpicture}
	\caption{Number of labels for each class among the two camera sets S1 and S2, and the two LiDAR sets S3 and S4.}
	\label{fig:num_labels_cam_lidar}
\end{figure}

The distribution of all labeled objects within the camera sets (S1 and S2) and LiDAR sets (S3 and S4) is visualized in Fig. \ref{fig:num_labels_cam_lidar}.
Cars account for the largest amount of labeled objects in all sets as they are most present on the freeway. 

On average the dataset contains 13.17 labels per frame. The average number of labeled objects per frame for each subset is displayed in the Table \ref{tbl:subsets}.

\section{Evaluation}

Here, we want to briefly present the results from an evaluation we carried out based on the A9 dataset. The purpose of this is to get a first estimate on possible performance improvements that can be accomplished with fine-tuning a detection model using our dataset. While training with large datasets such as MS COCO \cite{lin2015microsoft} allows for diverse object classifications, we want our perception system to be specialized on our specific application and sensor setup. 
Therefore, we want to validate whether a model that has been fine-tuned on our dataset indeed outperforms available pre-trained models.
In particular, we compare the detection performance of a YOLOv5 \cite{glenn_jocher_2021_5563715} model trained on the A9-Dataset from scratch (experiment E1) and the same model type trained on the MS COCO dataset before and after transfer learning on the A9-Dataset (experiments E2 and E3).

For this purpose, the A9-Dataset was split into a training (80\%), validation (10\%) and test set (10\%) using random indices. A RTX 3090 (with 24 GB of VRAM) GPU was used to train each model for 300 epochs using a batch size of 1. The evaluation was done on the same GPU using an input image size of 1152x1152 with a confidence threshold of 0.25 and intersection over union (IoU) threshold of 0.5.

The evaluation results of YOLOv5 on the A9 test set are shown in Table \ref{tbl:evaluation_results_yolov5}. 
Here, the average precision (AP) of the model for detecting objects of class car is given, as well as the inference time.
It follows from this table that the model from experiment 3 based on transfer learning (E3) achieves the best results in terms of AP.
For the model trained exclusively on the A9-Dataset (E1), the AP test results are $3.8\,\%$ higher than for the model initialized with the pre-trained MS COCO weights (E2).
Overall, a $4.2\,\%$ higher accuracy is achieved with YOLOv5-x-E3 (E3) for the car class compared to using only the MS COCO pre-trained weights.

\begin{table}[h]
	\caption{Evaluation results of experiments E1 to E3 for the car class on the A9 test set using an IoU value of 50\% and the mean of 10 IoU values between 50\% and 95\% in 5\% steps.}
	\label{tbl:evaluation_results_yolov5}
	\centering
	\begin{tabular}{|p{1.7cm}||p{1.6cm}||p{1.6cm}||p{1.7cm}|}
		\hline
		Model & $AP@0.5$ & $AP$ \newline $@[0.5:0.95]$ & Inference time \newline (FPS) \\
		\hline
		YOLOv5-x-E1 & 92.4 & 73.2 & 37 FPS\\
		\hline
		YOLOv5-x-E2 & 88.6 & 64.4 & 36 FPS\\
		\hline
		YOLOv5-x-E3 & 92.8 & 74.2 & 37 FPS \\
		\hline
	\end{tabular}
\end{table}

These results demonstrate the suitability of the A9-Dataset for fine-tuning existing perception models to meet the specific requirements of intelligent infrastructure systems, while at the same time also introducing additional object classes such as vans and trailers. 
\section{Conclusion}
\label{sec:conclusion}
The first version of the A9-Dataset from the Providentia++ test field containing LiDAR point clouds, camera images, and the associated 3D bounding boxes was presented. The sensor setup, the data and label formats were described and statistical information related to dataset contents was given. As part of future work the authors intend to release larger and more varied sets from the Providentia++ test field covering many more sensor types, traffic scenarios and challenging weather conditions.


\section*{Acknowledgment}

This work was funded by the Federal Ministry for Digital and Transport, Germany as part of the research project Providentia++. The authors would like to express their gratitude to the funding agency and to the numerous students at TUM who have contributed to the creation of the dataset.


\bibliographystyle{IEEEtran}


\end{document}